\documentclass[a4paper,fleqn]{cas-dc}

\usepackage[numbers]{natbib}

\usepackage{float}
\usepackage{multirow}
\usepackage{booktabs}
\usepackage{graphicx}  % for \resizebox
\usepackage{tabularx}
\def\tsc#1{\csdef{#1}{\textsc{\lowercase{#1}}\xspace}}
\tsc{WGM}
\tsc{QE}
\tsc{EP}
\tsc{PMS}
\tsc{BEC}
\tsc{DE}
\begin{document}
\let\WriteBookmarks\relax
\def\floatpagepagefraction{1}
\def\textpagefraction{.001}

% Short title
\shorttitle{Deep Learning for Soil Moisture Estimation}

% Short author
\shortauthors{A. Canovas-Rodriguez et~al.}

% Main title of the paper
\title [mode = title]{Deep Learning for Soil Moisture Estimation: Fusing Satellite Data with Optimally-Lagged Meteorological Features}                      
% direcciones a explorar:
%Downward infiltration — after rainfall, water enters the surface and percolates to deeper layers over hours to days. So shallow moisture leads deep moisture in time.
%Capillary rise — in dry conditions or with a high water table, moisture moves upward from deeper layers to the surface. So deep moisture leads shallow moisture.

% Title footnote mark
% eg: \tnotemark[1]
%\tnotemark[1,2]

% Title footnote 1.
% eg: \tnotetext[1]{Title footnote text}
% \tnotetext[<tnote number>]{<tnote text>} 

% Authors
\author[1]{Adrian Canovas-Rodriguez}[
    orcid=0009-0002-7986-5615
]
\ead{adriancr@um.es}

\author[1]{Aurora González Vidal}
\ead{aurora.gonzalez2@um.es}
% Corresponding author
\cormark[1]
\author[1]{Antonio F. Skarmeta}
\ead{skarmeta@um.es}

% Shared affiliation
\affiliation[1]{
    organization={Department of Information and Communication Engineering, University of Murcia},
    city={Murcia},
    country={Spain}
}
\cortext[1]{Corresponding author}

% For a title note without a number/mark
%\nonumnote{This note has no numbers. In this work we demonstrate $a_b the formation Y\_1 of a new type of polariton on the interfac between a cuprous oxide slab and a polystyrene micro-sphere placedon the slab.}

% Here goes the abstract
\begin{abstract}
Accurate soil moisture estimation in semi-arid agricultural regions requires integrating remote sensing and meteorological information while accounting for the delayed response of soil moisture to atmospheric forcing. This study introduces a Cross-Correlation Function (CCF) methodology to determine optimal temporal lags (0–30 days) between meteorological variables and soil moisture, as well as inter-depth lags (0–15 days) describing vertical moisture propagation from the surface (10 cm) to deeper layers (20–50 cm). The approach was validated across seven agricultural plots in southeastern Spain.

Three deep learning architectures, each targeting a distinct prediction granularity, were evaluated under five feature configurations ranging from satellite-only to full satellite--meteorology--depth fusion: a CNN for per-pixel estimation within each plot, an LSTM for frame-level (daily plot-mean) prediction, and a CNN-LSTM hybrid operating on sliding windows with pooled multi-patch training. Models were assessed on held-out data to measure genuine generalisation.

Meteorological variables improved performance over the satellite-only baseline, while subsurface depth information proved decisive across all architectures. The per-pixel CNN achieved the strongest single-patch result (R² = 0.877, RMSE = 2.28), with a seven-patch average R² of 0.535, representing an improvement of +1.00 over the satellite-only baseline. The pooled CNN-LSTM hybrid obtained the highest overall performance (R² = 0.930, CVRMSE = 8.0\%). These results demonstrate that explicitly modelling atmospheric and vertical subsurface delays substantially improves soil moisture estimation for precision agriculture.

\end{abstract}

\begin{highlights}
\item CCF identifies optimal meteorological (0--30 d) and inter-depth (0--15 d) lags
\item Per-pixel CNN evaluated with a date-grouped split to avoid data leakage
\item Subsurface depth memory is decisive for held-out soil moisture prediction
\item Depth features raise the held-out CNN average R² from satellite-only (-0.47) to full fusion (0.535)
\item Pooled CNN-LSTM hybrid attains the best held-out score (R2=0.93, CVRMSE=8\%)
\end{highlights}

% Keywords
% Each keyword is seperated by \sep

\maketitle

\section{Introduction}
\label{sec:intro}

Soil moisture is a critical state variable of the hydrological cycle, governing evapotranspiration, runoff generation and solute transport across agricultural landscapes, and its dynamics in semi-arid climates present unique challenges owing to their high spatiotemporal variability~\cite{manrique2017soil}. Soil moisture influences crop conditions, affecting evapotranspiration rates, which subsequently impacts runoff, particularly over large agricultural areas~\cite{ines2013assimilation}. At finer scales, it plays a vital role in biogeochemical processes such as solute transport, directly influencing water quality~\cite{ayers1985water}.

The evolution of soil moisture measurement methods---from early gravimetric techniques to electrical and, more recently, advanced electronic sensing---has improved monitoring accuracy. Traditional approaches included gravimetric sampling, where soil samples were oven-dried to assess moisture content. The introduction of electrical techniques, such as tensiometers and Time-Domain Reflectometry (TDR), marked a significant advancement by providing continuous data~\cite{robinson2003review}. Today, Internet of Things (IoT)-based sensing allows real-time, large-scale monitoring, enabling farmers to manage agricultural areas remotely~\cite{farooq2019survey, garg2016application}. However, while soil probes offer high precision at specific points, their spatial representation is limited as the distance from the measurement point increases. This highlights the need for supplementary, scalable methods such as remote sensing.

Satellite remote sensing provides spatially continuous observations across several sensing domains. Spaceborne microwave missions such as SMOS~\cite{kerr2010smos} and SMAP~\cite{entekhabi2010soil} deliver global surface soil moisture at near-daily revisit but at coarse spatial resolutions of 25--40~km, insufficient for field-scale agriculture, while Synthetic Aperture Radar offers higher resolution at the cost of strong assumptions on surface roughness and vegetation. In the optical domain, Sentinel-2 offers 10~m multispectral imagery with a 5-day revisit cycle, well suited to large-scale agricultural monitoring~\cite{drusch2012sentinel}: its multispectral and shortwave-infrared bands enable spectral indices---such as NDMI, NDVI and SAVI---that act as indirect proxies for surface moisture through vegetation vigour and bare-soil reflectance. Even so, optical retrievals struggle to infer moisture beyond the shallow surface layer~\cite{njoku2003soil}.

Recent advances in deep learning have enabled powerful data-driven approaches for soil moisture estimation, and recent reviews report that learning-based methods now dominate over empirical and physical models~\cite{wang2023deep}. Convolutional Neural Networks (CNNs) excel at extracting spatial features from pixel-level data~\cite{hegazi2023prediction}, while Long Short-Term Memory (LSTM) networks are well-suited for temporal sequence modelling~\cite{geng2024enhancing}. Hybrid architectures that combine both paradigms have shown promise in capturing joint spatial--temporal dependencies~\cite{yu2021hybrid}. Several studies have applied machine learning to soil moisture estimation using remote sensing~\cite{rani2022machine, ahmad2010estimating}, while others have explored deep learning for agricultural prediction tasks~\cite{lee2019estimation}. However, most studies that fuse satellite and meteorological data for soil moisture prediction employ same-day meteorological variables, neglecting the temporal delay between atmospheric forcing and its propagation into the soil profile.

This temporal delay reflects a fundamental property of soil moisture: \emph{memory}. The present moisture state retains information about past forcing over time scales from days to months, modulated by soil texture, vegetation and antecedent wetness~\cite{wei2025spatiotemporal}, and multi-day accumulation windows of meteorological forcing are consistently more informative than instantaneous values---14-day antecedent precipitation, for example, is far more predictive than sub-daily totals~\cite{bai2025near}. Memory is also expressed vertically: surface moisture changes propagate downward through the unsaturated zone with depth-increasing delays, so deeper layers carry a lagged, smoothed record of surface dynamics. A substantial literature estimates root-zone or profile soil moisture \emph{from} the surface~\cite{rahmati2024soil}, using data-driven models including ConvLSTM~\cite{wu2025integrating}, almost always treating the deeper layers as the prediction \emph{target}. The complementary direction---exploiting the deeper layers as lagged predictive \emph{memory} for the surface estimate---is rarely exploited as additional predictive information.

A further, often overlooked issue concerns how such models are evaluated. The conventional random train--test split is invalid when observations are not independent, as is the case for spatially and temporally autocorrelated environmental data; dependence between training and test samples inflates reported accuracy and is a principal driver of the reproducibility problems in machine-learning-based science~\cite{kapoor2023leakage}. In geospatial modelling, ignoring data structure during cross-validation yields over-optimistic estimates and masks overfitting, and dependence-aware schemes (spatial blocking, leave-location-out, temporally blocked validation) are required for honest assessment~\cite{roberts2017cross}; the same hazard motivates leave-profile-out validation in digital soil mapping~\cite{le2014spatial}. The concern is acute for per-pixel soil-moisture models trained against a point sensor: a single daily ground-truth value is shared by every pixel of a plot on a given date, so a naive pixel-level random split places near-identical spectra carrying the same target on both sides of the partition, and the resulting metric rewards within-scene reconstruction rather than genuine temporal generalisation. This form of group leakage is rarely addressed in the remote-sensing soil-moisture literature.

This study addresses these gaps by introducing a Cross-Correlation Function (CCF) analysis to determine: (i)~the optimal temporal offset between each meteorological variable and soil moisture, and (ii)~the optimal temporal lag between surface (10~cm) and deeper soil moisture layers (20--50~cm). The approach was validated over seven agricultural plots in the Region of Murcia (southeastern Spain), as illustrated in Figure~\ref{fig:map}. The main contributions of this work include:

\begin{enumerate}
    \item A CCF-based methodology to identify optimal temporal lags (0--30~days) between five meteorological variables and soil moisture, and inter-depth lags (0--15~days) between surface and subsurface soil moisture layers, providing physically interpretable feature engineering for deep learning models.
    \item A temporally-aligned feature-level fusion scheme that concatenates Sentinel-2 spectral features, agrometeorological variables, and multi-depth in-situ soil moisture into a single input vector but---unlike the conventional lag-$0$ fusion adopted in most multi-source studies---shifts each meteorological and subsurface stream by its own CCF-optimal lag before concatenation, adapting the temporal registration of the fused vector to local soil behaviour.
    \item Evaluation of three deep learning architectures---a per-pixel CNN, a per-date LSTM, and a windowed pooled CNN-LSTM hybrid---each targeting a different prediction granularity, under five feature configurations: satellite-only, satellite with same-day (lag~$= 0$) meteorological data, satellite with optimally-lagged meteorological data, and two configurations incorporating depth-lagged subsurface soil moisture (same-day and optimally-lagged). All architectures are evaluated on held-out data; in particular the per-pixel CNN is trained under a date-grouped split (\texttt{GroupShuffleSplit} on the acquisition date), so that no acquisition date appears on both sides of the partition and its metric measures genuine held-out-date generalisation directly comparable to the LSTM's, rather than within-date reconstruction. Because the architectures solve different prediction tasks on datasets differing in size by orders of magnitude (Section~\ref{subsec:comparison}), performance is reported within each architecture family rather than as a single cross-architecture benchmark. The same-day meteorological baseline isolates the specific benefit of CCF-based lag optimisation over the conventional approach used in most prior studies.
    \item A comprehensive data processing pipeline that integrates ground-truth sensor measurements, Sentinel-2 multispectral imagery, daily agrometeorological observations from the SIAR network, and multi-depth soil moisture profiles, together with a per-pixel raster product that projects the plot-level estimate to Sentinel-2 resolution.
    \item A CNN-LSTM hybrid architecture that uses sliding-window sequences over pooled multi-patch data with one-hot patch encoding, enabling joint learning of spatial and temporal patterns.
\end{enumerate}

The remainder of this paper is organised as follows. Section~\ref{sec:materials} describes the study area, data sources, and deep learning methodologies. Section~\ref{sec:ed} presents the experimental design, including the multi-source feature fusion strategy, the data collection and processing pipeline, dataset construction, the five model configurations, and the model training protocol. Section~\ref{sec:results} reports and analyses the experimental results. Section~\ref{sec:discussion} discusses the findings, and Section~\ref{sec:conclusion} presents conclusions and future directions.

\section{Materials and Methods}
\label{sec:materials}

\subsection{Study Area and Experimental Plots}
\label{subsec:study_area}

The study was conducted in the Region of Murcia, southeastern Spain (approximately 37\textdegree\,55$'$\,N, 1\textdegree\,28$'$\,W), a semi-arid Mediterranean environment with hot summers, mild winters, a mean annual temperature near 18\textdegree C, and scarce, irregular rainfall (300--350~mm\,yr$^{-1}$, concentrated between October and March). These conditions produce high spatiotemporal variability in soil moisture and make the region representative of water-limited agriculture across the Mediterranean basin.

\begin{figure}[pos=H]
    \centering
    \includegraphics[width=\linewidth]{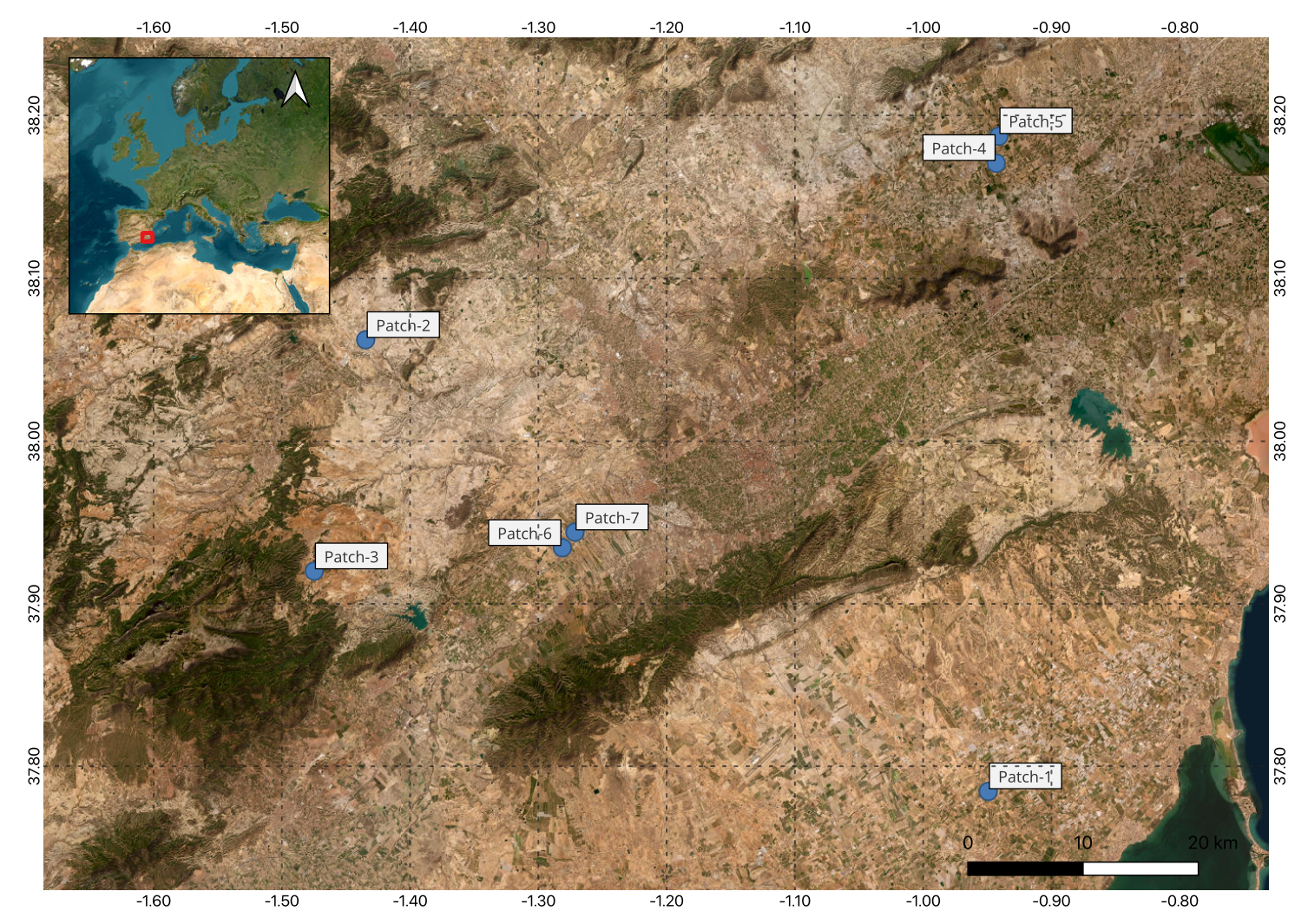}
    \caption{Location of the seven study plots on the orthophoto of the Region of Murcia.}
    \label{fig:map}
\end{figure}

\begin{figure*}[ht]
    \centering
    \includegraphics[width=\linewidth]{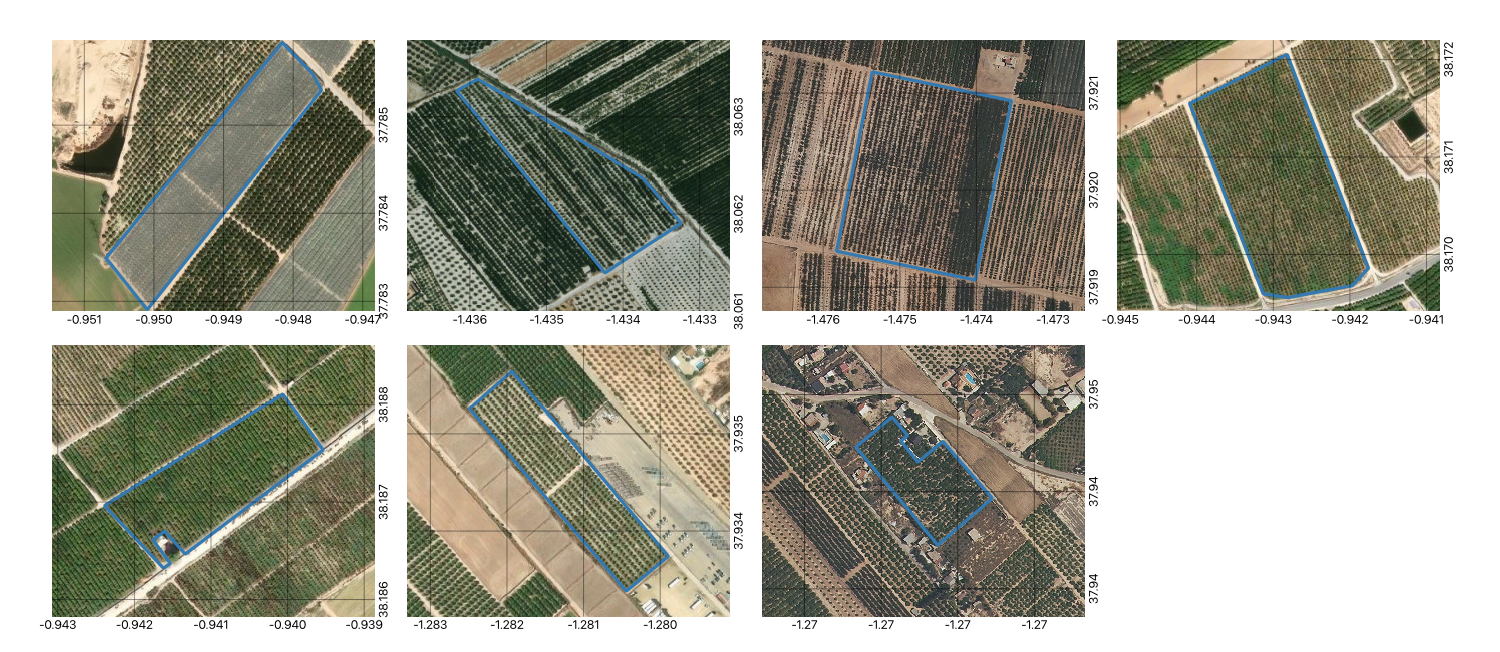}
    \caption{Aerial views of the seven study plots (Patch-1 to Patch-7) located in the Region of Murcia, Spain.}
    \label{fig:patches}
\end{figure*}

Seven rainfed and irrigated plots (hereafter Patch-1 to Patch-7), distributed across the La Palmera, Torre Pacheco and ``PSB Producci'{o}n Vegetal'' estates, were instrumented for this study (Figure~\ref{fig:map}). The plots span contrasting crop types, irrigation practices and soil characteristics (Table~\ref{tab:plots}). Representative views of the monitored plots are shown in Figure~\ref{fig:patches}. Together, these sites provide the diversity in soil moisture dynamics needed to assess model generalisation across agronomic settings.

\begin{table}[H]
\centering
\caption{Characteristics of the seven study plots. Acquisition dates are the number of cloud-free Sentinel-2 frames available per plot.}
\label{tab:plots}
\resizebox{\columnwidth}{!}{%
\begin{tabular}{llllc}
\toprule
\textbf{Plot} & \textbf{Estate} & \textbf{Crop} & \textbf{SIAR station} & \textbf{\# dates} \\
\midrule
Patch-1 & Torre Pacheco          & Citrus & TP73 & 104 \\
Patch-2 & La Palmera             & Peach & ML21 & 125 \\
Patch-3 & La Palmera             & Peach & ML21 & 76  \\
Patch-4 & PSB Producci\'{o}n Veg.& Peach & MO41 & 95  \\
Patch-5 & PSB Producci\'{o}n Veg.& Peach & MO41 & 119 \\
Patch-6 & Sierra Espu\~{n}a area & Citrus & MU31 & 202 \\
Patch-7 & Sierra Espu\~{n}a area & Citrus & MU31 & 25  \\
\bottomrule
\end{tabular}
}
\end{table}

\subsection{Soil Moisture Sensors}
\label{subsec:sensors}

Ground-truth soil moisture was measured with Sentek Drill~\&~Drop capacitance probes, which record volumetric water content at multiple depths every 30~minutes with a manufacturer-specified accuracy of $\pm 0.003$ (m\textsuperscript{3}\,m\textsuperscript{-3}). All plots provided consistent readings over the top five depths (10, 20, 30, 40 and 50~cm), which define the profile used throughout. The prediction target is the daily-mean volumetric soil moisture at 10~cm; the deeper layers (20--50~cm) are used as candidate predictive features through the inter-depth analysis of Section~\ref{subsubsec:depth_ccf}.

\subsection{Satellite Imagery}
\label{subsec:satellite}

Multispectral imagery was obtained from the twin Sentinel-2 satellites of ESA's Copernicus programme, which provide a five-day revisit at the equator and 13 spectral bands (443--2190~nm) at 10, 20 and 60~m resolution. Imagery was acquired through the SentinelHub Python API, downloaded as GeoTIFF, clipped to each plot with GeoJSON boundaries, masked to the plot interior, and atmospherically corrected to Level-2A surface reflectance. The cirrus band (B10), which carries no surface information, was discarded, leaving 12 bands (B01--B09, B8A, B11, B12) as inputs.

\subsection{Satellite-Based Spectral Indices}
\label{subsec:indices}

Twelve spectral indices spanning vegetation vigour, canopy moisture and bare-soil response were derived from the Sentinel-2 bands (Table~\ref{tab:indices}). Together with the 12 bands, they constitute the 24 satellite-derived features supplied to the models.

\begin{table}[H]
\centering
\caption{Spectral indices derived from Sentinel-2 bands used as input features.}
\label{tab:indices}
\resizebox{\columnwidth}{!}{%
\begin{tabular}{lcp{5cm}}
    \toprule
    \textbf{Index} & \textbf{Bands} & \textbf{Description} \\
    \midrule
    NDVI  & B4, B8      & Normalised Difference Vegetation Index \\
    ARVI  & B2, B4, B8  & Atmospherically Resistant Vegetation Index \\
    GCI   & B3, B8      & Green Chlorophyll Index \\
    GNDVI & B3, B8      & Green NDVI \\
    NBR2  & B11, B12    & Normalised Burn Ratio 2 \\
    NDBI  & B8, B11     & Normalised Difference Built-up Index \\
    NDMI  & B8, B11     & Normalised Difference Moisture Index \\
    NDRE  & B5, B8      & Normalised Difference Red Edge \\
    NDSI  & B3, B11     & Normalised Difference Snow Index \\
    NDWI  & B3, B8      & Normalised Difference Water Index \\
    REIP  & B5, B8      & Red-edge reflectance ratio (B8/B5) \\
    SAVI  & B4, B8      & Soil-Adjusted Vegetation Index \\
    \bottomrule
\end{tabular}
}
\end{table}

\subsection{Meteorological Data}
\label{subsec:meteo}

Daily agrometeorological data were obtained from the SIAR network. Rather than a single regional station, each plot was assigned its nearest SIAR station with complete coverage of the target variables over the measurement period, giving four stations across the seven plots (Table~\ref{tab:plots}). Five variables with established control on soil moisture were used: mean, maximum and minimum air temperature (\textdegree C), daily accumulated precipitation (mm), and daily accumulated solar radiation (MJ\,m\textsuperscript{-2}). Temperature governs evapotranspiration, precipitation supplies water, and radiation drives the surface energy balance. Because these effects are not instantaneous---rainfall infiltrates over hours to days and evapotranspiration accumulates over several---the optimal temporal alignment between each variable and soil moisture is determined empirically in Section~\ref{subsec:ccf}.

\subsection{Cross-Correlation Function for Optimal Lag Estimation}
\label{subsec:ccf}

A central methodological contribution is the use of the Cross-Correlation Function (CCF) to identify the optimal temporal offset between each meteorological variable and soil moisture. For each plot $p$ and variable $v$, the Pearson correlation between the daily soil moisture series $\mathrm{SM}_t$ and the variable shifted by $d$ days is
\begin{equation}
    r(d) = \frac{\mathrm{cov}(\mathrm{SM}_t,\; v_{t-d})}{\sigma_{\mathrm{SM}}\,\sigma_v},
    \qquad d \in \{0, 1, \ldots, 30\},
    \label{eq:ccf}
\end{equation}
and the optimal lag is the shift maximising the absolute correlation,
\begin{equation}
    d^* = \arg\max_{d \in [0,\,30]} \lvert r(d) \rvert .
    \label{eq:optlag}
\end{equation}
Computing $d^*$ independently per variable and plot adapts the feature engineering to local soil--atmosphere dynamics. The CCF captures linear associations; nonlinear couplings are left to the subsequent learning models and are not resolved by the lag search itself.

\subsubsection{Inter-Depth CCF for Vertical Soil Moisture Propagation}
\label{subsubsec:depth_ccf}

An analogous inter-depth CCF models the vertical propagation of moisture. For each plot and deeper layer $z \in \{20, 30, 40, 50\}$~cm, the correlation between the surface (10~cm) moisture and the lagged deeper-layer moisture is
\begin{equation}
    r_z(d) = \frac{\mathrm{cov}(\mathrm{SM}_{10\mathrm{cm},\,t},\; \mathrm{SM}_{z,\,t-d})}
                  {\sigma_{\mathrm{SM}_{10\mathrm{cm}}}\,\sigma_{\mathrm{SM}_z}},
    \qquad d \in \{0, 1, \ldots, 15\},
    \label{eq:depth_ccf}
\end{equation}
with optimal lag $d^*_z = \arg\max_{d \in [0,\,15]} \lvert r_z(d) \rvert$. The deeper-layer values, shifted by $d^*_z$, supply the models with subsurface ``memory'' that optical reflectance cannot observe.

\begin{equation}
    d^*_z = \arg\max_{d \in [0,\,15]} \left| r_z(d) \right|
    \label{eq:depth_optlag}
\end{equation}

The optimal inter-depth lag $d^*_z$ is defined analogously to Equation~\ref{eq:optlag}, as the shift maximising the absolute correlation between the surface and each deeper layer.

\subsection{Deep Learning Architectures}
\label{subsec:models}

Three architectures were implemented, each targeting a different prediction granularity (per-pixel, per-date, and windowed multi-plot). Hyperparameters were fixed across feature configurations to isolate the effect of the inputs; the training and evaluation protocol is detailed in Section~\ref{sec:ed}.

\subsubsection{Convolutional Neural Network (CNN)}
\label{subsubsec:cnn}

The per-pixel CNN~\cite{shah2025systematic} comprises four Conv1D blocks with increasing filter counts (64, 128, 256, 512), each with batch normalisation; the first three blocks are additionally followed by max-pooling and dropout (0.25), while the fourth feeds directly into a flattening layer. The flattened output feeds two dense layers (256 and 64 units; dropout 0.4 and 0.2) and a single-neuron regression head. Training uses Adam with MSE loss for up to 200 epochs (batch size 64, early stopping patience 15). Features are scaled with a RobustScaler, chosen because the per-pixel reflectance distribution contains outliers from residual atmospheric and sensor noise.

\subsubsection{Long Short-Term Memory Network (LSTM)}
\label{subsubsec:lstm}

The frame-level LSTM~\cite{hochreiter1997lstm} consists of a stack of recurrent layers comprising a bidirectional LSTM layer with 512 units, followed by an alternating tower of GRU (256), LSTM (128), GRU (64) and LSTM (32) layers. A final dense layer with a single neuron produces the soil moisture estimate. Each recurrent layer is followed by dropout (0.4) and batch normalisation, with inputs shaped $(1, F)$. Training uses Adam (lr~0.001) and MSE loss for up to 900 epochs (batch size 64, early stopping patience 20). A MinMaxScaler is applied to features and target, and predictions are inverse-transformed before scoring; min--max scaling is preferred here because the small frame-level samples lack the outlier mass that motivates robust scaling for the CNN.

\subsubsection{CNN-LSTM Hybrid}
\label{subsubsec:hybrid}

The hybrid~\cite{kandamali2025hybrid} operates on sliding windows of $W = 10$ consecutive dates (stride 1). Two Conv1D layers (64 and 128 filters, kernel 3, ``same'' padding; batch normalisation, dropout 0.2) extract local temporal features; two stacked LSTM layers (128 and 64 units; dropout 0.3) model longer-range dependence; and a Dense(64, ReLU) layer with dropout (0.2) precedes the regression head. All seven plots are pooled, with plot identity encoded as a seven-dimensional one-hot vector, so the model shares temporal dynamics across plots while remaining plot-aware. Training uses Adam (lr~0.001) and MSE loss for up to 300 epochs (batch size 32, early stopping patience 20), with a MinMaxScaler applied to all but the one-hot features.

\subsection{Evaluation Metrics}
\label{subsec:metrics}

Model performance is reported with five complementary metrics, where $y_i$ and $\hat{y}_i$ are the observed and predicted values, $\bar{y}$ the observed mean, and $n$ the number of test samples:

\begin{equation}
\mathrm{RMSE} = \sqrt{\tfrac{1}{n}\sum_{i=1}^{n}(y_i-\hat{y}_i)^2}, \quad
\mathrm{MAE} = \tfrac{1}{n}\sum_{i=1}^{n}\lvert y_i-\hat{y}_i\rvert
\end{equation}

\begin{equation}
\mathrm{CVRMSE} = \tfrac{\mathrm{RMSE}}{\bar{y}}\times 100\%, \quad
R^2 = 1-\frac{\sum_i (y_i-\hat{y}_i)^2}{\sum_i (y_i-\bar{y})^2}
\end{equation}

and the Pearson correlation is defined as
\begin{equation}
\mathrm{CC}=\frac{\mathrm{cov}(y,\hat{y})}{\sigma_y\sigma_{\hat{y}}}.
\end{equation}

CVRMSE normalises the error by mean soil moisture, enabling comparison across plots with different baseline levels.

\section{Experimental Design}\label{sec:ed}

\begin{figure*}[hbtp]
    \centering
    \includegraphics[width=0.65\linewidth]{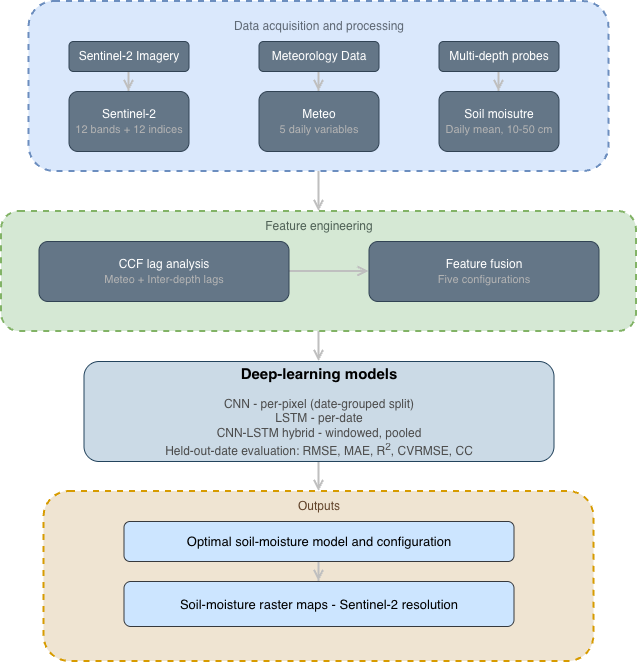}
    \caption{Overall processing workflow from Sentinel-2 and meteorological inputs through feature engineering (including CCF-based lag estimation) to deep learning models and evaluation.}
    \label{fig:pipeline}
\end{figure*}

Figure~\ref{fig:pipeline} summarises the feature fusion strategy, where satellite-derived features are combined with meteorological variables and deep soil moisture observations under different temporal alignment schemes. Depending on the configuration, external variables are either used at the same-day timestamp or shifted using CCF-derived optimal lags before being merged with satellite inputs to form the final model feature vectors.

\subsection{Multi-Source Feature Fusion Strategy}
\label{subsec:fusion}

The central modelling problem of this study is how to combine three heterogeneous information sources---Sentinel-2 spectral observations, agrometeorological records, and multi-depth in-situ soil moisture---into a single estimator of surface (10~cm) soil moisture. In the taxonomy of multi-source remote-sensing fusion, integration can occur at three stages~\cite{li2022deep}: \emph{decision-level} (late) fusion, where a separate model is trained per source and the outputs are combined into an ensemble; \emph{intermediate} fusion, where each source is encoded by a dedicated network branch and the learned representations are concatenated; and \emph{feature-level} (early) fusion, where the features of all sources are concatenated into a single input vector that one model learns jointly. Across a survey of more than 950 studies, feature-level fusion is the most widely adopted strategy for multi-sensor remote sensing~\cite{li2022deep}.

This work adopts feature-level fusion: for each acquisition date, the satellite features (bands and indices), the meteorological variables, and the deeper-layer soil moisture are concatenated into a single vector that one network maps to the 10~cm target. The choice is deliberate. The prediction target is one daily value from a point probe, shared by every modality on a given date, so the three sources are already co-registered in time and require no per-source calibration; a single network can then learn the cross-modal interactions directly. Equally important, the frame-level datasets are small (20--160 observations per plot; Table~\ref{tab:dataset_sizes}), which makes decision-level ensembling or multi-branch intermediate fusion---each of which multiplies the number of free parameters and the per-source data demand---impractical to fit reliably; feature-level fusion concentrates the parameter budget on a single model.

The novel element of the proposed fusion is therefore \emph{not} the concatenation mechanism, which is standard, but the \emph{temporal alignment} applied before concatenation. Most studies that fuse satellite with auxiliary data merge every source at the same acquisition date (lag~$= 0$). Instead, we shift each meteorological variable and each deeper soil-moisture layer by its own CCF-optimal lag (Section~\ref{subsec:ccf}) before merging. The motivation is physical and soil-specific: whether---and after how many days---a near-surface signal observable from orbit is reflected in the probe reading depends on the soil itself. Permeability, porosity, and water-retention capacity govern how fast infiltrating water percolates through the profile and how long moisture is held near the surface, so the delay between atmospheric forcing (or a deeper-layer state) and the 10~cm response is necessarily plot-specific. Fusing every source at lag~0 implicitly assumes an instantaneous, soil-independent coupling that the data contradict: the inter-depth CCF (Section~\ref{subsubsec:depth_ccf}) recovers near-instantaneous surface--subsurface coupling on a well-connected plot (Patch-3, $r \approx 0.95$ to 50~cm) but 14--15-day lags and weak correlation on a plot with a distinct, poorly connected vertical profile (Patch-1, $|r| < 0.26$). Aligning each stream by its empirically estimated lag lets the fused vector carry the moisture state that is actually predictive of the target, adapting the fusion to local soil behaviour rather than imposing a fixed temporal registration.

Finally, the CNN-LSTM hybrid adds an \emph{architectural} fusion of spatial and temporal information---its convolutional layers extract local temporal--spectral structure within a sliding window before the recurrent layers model longer-range dynamics---whereas the standalone CNN and LSTM fuse the sources only at the input. The five feature configurations defined in Section~\ref{subsec:configs} are concrete instances of this strategy, differing in which sources are fused and under which temporal alignment.

\subsection{Data Preprocessing}
\textit{Soil Moisture Data Preprocessing}

Field measurement of soil moisture is subject to multiple difficulties including sensor failures, power outages, and communication losses.
Missing values---representing unavailable information at a given time---arise because the sensor cannot perform the measurement, the datalogger memory becomes full, or the sensor temporarily loses power.
Since physical conditions change over time, missing measurements cannot be recovered retrospectively.

\begin{figure}[pos = H]
    \centering
    \includegraphics[width=0.85\linewidth]{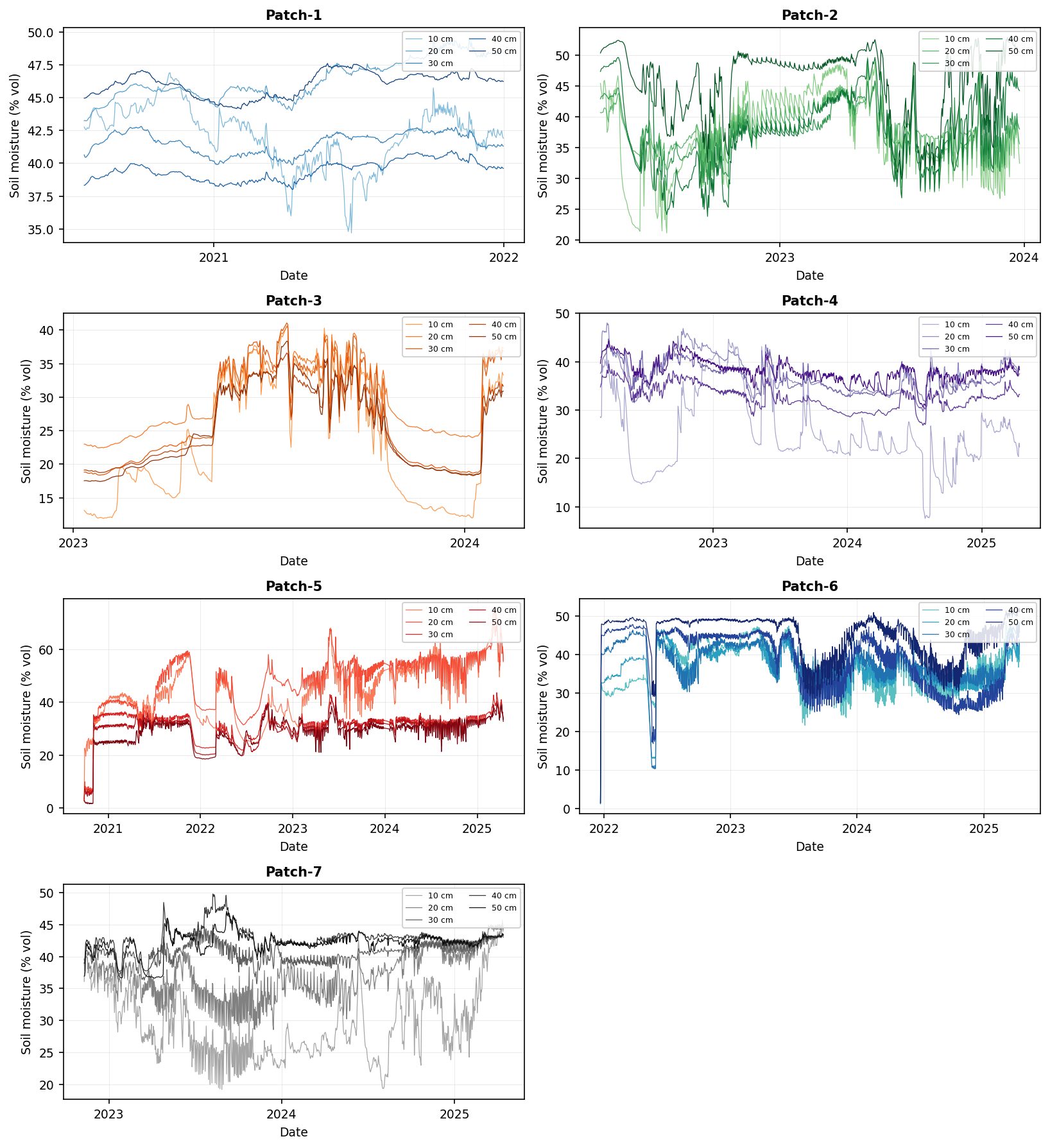}
    \caption{Temporal evolution of soil moisture measured by depth sensors across the study plots. Each line represents a different measurement depth (10--50~cm).}
    \label{fig:datoshumedad}
\end{figure}

The Sentek Drill~\&~Drop sensors installed across the seven study plots recorded soil moisture at 30-minute intervals over the five depths (10--50~cm) used in this study.
Figure~\ref{fig:datoshumedad} illustrates the temporal evolution of soil moisture across measurement depths.
A data cleaning pipeline was applied, including outlier detection, gap identification, and linear interpolation for missing values.
The target variable used for model training is the daily average of all 30-minute measurements.

\textit{Satellite Image Acquisition and Processing}

Sentinel-2 imagery was acquired via the SentinelHub library, which interfaces with the Copernicus API through the \texttt{SHConfig()} authentication mechanism.
For each acquisition date and plot, a GeoTIFF raster containing all spectral bands was downloaded, clipped to the area of interest using GeoJSON boundary files, and masked to exclude pixels outside the plot boundaries.

From each satellite image, 12 Sentinel-2 spectral bands (B01--B09, B8A, B11, B12; the cirrus band B10 is discarded) and 12 vegetation/soil indices (Table~\ref{tab:indices}) were computed.
Two dataset formats were constructed:

\begin{itemize}
    \item \textbf{Frame-level dataset (for LSTM)}: For each acquisition date, the 12 bands and 12 indices are summarised across all valid (non-masked) pixels using minimum, mean, and maximum statistics, yielding a single observation per date with $24 \times 3 = 72$ features.
    Dataset sizes across the seven patches are shown in Table~\ref{tab:dataset_sizes}.

    \item \textbf{Pixel-level dataset (for CNN)}: Each pixel of each image constitutes an independent observation with the raw band and index values (24 features), along with spatial coordinates.
    Dataset sizes range from approximately 2{,}450 (Patch-7) to 30{,}000 (Patch-4) observations per patch.
\end{itemize}

\begin{table}[H]
\centering
\caption{Dataset sizes for each patch. Frame-level datasets (LSTM) contain one row per acquisition date; pixel-level datasets (CNN) contain one row per pixel per date.}
\label{tab:dataset_sizes}
\begin{tabular}{lrr}
    \toprule
    \textbf{Patch} & \textbf{Frame-level (LSTM)} & \textbf{Pixel-level (CNN)} \\
    \midrule
    Patch-1 & 104 & 27{,}976 \\
    Patch-2 & 125 & 25{,}125 \\
    Patch-3 &  76 & 29{,}640 \\
    Patch-4 &  95 & 30{,}020 \\
    Patch-5 & 119 & 24{,}038 \\
    Patch-6 & 202 & 14{,}740 \\
    Patch-7 &  25 &  2{,}450 \\
    \bottomrule
\end{tabular}
\end{table}

\textbf{Meteorological Data Processing}

Daily meteorological records from each plot's assigned SIAR station were parsed from the SIAR data format, converting Spanish decimal notation (comma as decimal separator) to standard floating-point values.
Records with missing values in any of the five target variables were excluded.
The resulting clean dataset spans the full study period with daily resolution.

\subsection{Feature Configurations}\label{subsec:configs}

The five configurations below are concrete instances of the feature-level fusion strategy of Section~\ref{subsec:fusion}, progressively adding sources and tightening their temporal alignment so that the contribution of each fusion ingredient can be isolated. Each deep learning architecture was evaluated under all five:

\begin{enumerate}
    \item \textbf{sat\_only}: Satellite-derived features only (spectral bands and indices).
    This serves as the baseline, representing model performance without meteorological or depth augmentation.

    \item \textbf{sat+meteo\_lag0}: Satellite features augmented with same-day (lag~$= 0$) meteorological variables.
    No temporal offset is applied; this represents the conventional approach used in most prior studies and serves as a control to isolate the benefit of CCF-based lag optimisation.

    \item \textbf{sat+meteo\_optlag}: Satellite features augmented with optimally-lagged meteorological variables.
    Each meteorological variable is shifted by its CCF-optimal lag $d^*$ (Equation~\ref{eq:optlag}) before merging with the satellite data.

    \item \textbf{sat+meteo+depth\_lag0}: Satellite features augmented with optimally-lagged meteorological variables and same-day (lag~$= 0$) deep soil moisture from the 20--50~cm layers.

    \item \textbf{sat+meteo+depth\_optlag}: Satellite features augmented with optimally-lagged meteorological variables and optimally-lagged deep soil moisture, where each deeper layer is shifted by its inter-depth CCF-optimal lag $d^*_z$ (Equation~\ref{eq:depth_optlag}).
\end{enumerate}

The feature fusion is performed by date-based merging: for a satellite observation at date $t$, the meteorological variable $v$ is taken from date $t - d^*_v$ (or date $t$ for the lag~0 configuration), and the deep soil moisture at depth $z$ is taken from date $t - d^*_z$.
Observations with missing data after merging are excluded.
The resulting feature counts are: LSTM datasets have 72 (sat\_only), 77 (+meteo\_lag0 or +meteo\_optlag), or 81 (+meteo+depth) features; CNN datasets have 24, 29, or 33 features respectively.

\subsection{Model Training Protocol}

Table~\ref{tab:training_config} summarises the training configuration for each architecture.
All models use an 80/20 train--test split with the same random seed.
Because each acquisition date carries a single ground-truth sensor reading shared by all of its pixels, the CNN splits by \emph{acquisition date} (\texttt{GroupShuffleSplit}), so that no date---and therefore no shared per-date target---appears in both the training and test sets; the LSTM likewise splits at the date level, leaving entire dates unseen at test time; the CNN-LSTM hybrid splits at the \emph{window} level on the pooled set, so overlapping windows sharing dates may straddle the split.
All three architectures therefore report held-out generalisation, although at different sampling units and on training sets that differ by orders of magnitude, as discussed in Section~\ref{subsec:comparison}.
Early stopping on validation loss is employed to prevent overfitting and determine the effective number of training epochs.

\begin{table}[H]
\centering
\caption{Training configuration for each deep learning architecture.}
\label{tab:training_config}
\resizebox{\columnwidth}{!}{%
\begin{tabular}{lccc}
    \toprule
    \textbf{Parameter} & \textbf{CNN} & \textbf{LSTM} & \textbf{CNN-LSTM Hybrid} \\
    \midrule
    Max epochs          & 200   & 900   & 300 \\
    Batch size          & 64    & 64    & 32 \\
    Learning rate       & Adam default & 0.001 & 0.001 \\
    Early stop patience & 15    & 20    & 20 \\
    Feature scaler      & RobustScaler & MinMaxScaler & MinMaxScaler \\
    Target scaler       & None  & MinMaxScaler & MinMaxScaler \\
    Input shape         & $(F, 1)$ & $(1, F)$ & $(W, F)$, $W{=}10$ \\
    Training scope      & Per-patch & Per-patch & Pooled (all patches) \\
    Random seed         & 143   & 143   & 143 \\
    \bottomrule
\end{tabular}
}
\end{table}

\subsection{CNN-LSTM Sliding-Window Construction}

For the CNN-LSTM hybrid, the frame-level time series from each patch are sorted chronologically and converted into sliding windows of length $W = 10$ with stride 1.
Patch identity is encoded as a seven-dimensional one-hot vector appended to each observation's feature vector (one component per study plot).
All patches are pooled into a single training set, enabling the model to learn shared temporal dynamics while distinguishing between patches through the one-hot encoding.
The target value for each window is the 10~cm soil moisture measurement on the final (most recent) date of the window.

\section{Results}\label{sec:results}
\subsection{Cross-Correlation Analysis}
\label{subsec:ccf_results}

The CCF analysis (Equations~\ref{eq:ccf}--\ref{eq:optlag}) was applied to each patch--variable combination across lags from 0 to 30 days.
Table~\ref{tab:ccf_results} presents the optimal lag and corresponding Pearson correlation for each combination.

\begin{table}[H]
\centering
\caption{Optimal lag (days) and Pearson correlation $|r|$ between soil moisture and meteorological variables, determined by CCF analysis for each of the seven patches (each using its assigned SIAR station).}
\label{tab:ccf_results}
\resizebox{\columnwidth}{!}{%
\begin{tabular}{lllcc}
    \toprule
    \textbf{Patch} & \textbf{Station} & \textbf{Variable} & \textbf{Optimal Lag (days)} & \textbf{$|r|$ at Lag} \\
    \midrule
\multirow{5}{*}{Patch-1} & \multirow{5}{*}{TP73}
    & TempMean & 30 & 0.266 \\
& & TempMax & 30 & 0.301 \\
& & TempMin & 28 & 0.259 \\
& & Precipitation & 30 & 0.302 \\
& &  Radiation & 1 & 0.483 \\
    \midrule
    \multirow{5}{*}{Patch-2} & \multirow{5}{*}{ML21}
        & TempMean & 3 & 0.559 \\
      &  & TempMax & 1 & 0.533 \\
      &  & TempMin & 8 & 0.570 \\
     &   & Precipitation & 0 & 0.255 \\
    &    & Radiation & 29 & 0.550 \\
    \midrule
    \multirow{5}{*}{Patch-3} & \multirow{5}{*}{ML21}
        & TempMean & 0 & 0.786 \\
    &    & TempMax & 0 & 0.746 \\
    &    & TempMin & 0 & 0.804 \\
    &    & Precipitation & 28 & 0.293 \\
    &    & Radiation & 27 & 0.722 \\
    \midrule
    \multirow{5}{*}{Patch-4} & \multirow{5}{*}{MO41}
        & TempMean & 11 & 0.378 \\
   &     & TempMax & 11 & 0.380 \\
  &      & TempMin & 11 & 0.374 \\
  &      & Precipitation & 3 & 0.318 \\
 &       & Radiation & 7 & 0.220 \\
    \midrule
    \multirow{5}{*}{Patch-5} & \multirow{5}{*}{MO41}
        & TempMean & 30 & 0.395 \\
&       & TempMax & 27 & 0.393 \\
  &      & TempMin & 30 & 0.398 \\
  &      & Precipitation & 30 & 0.200 \\
 &      & Radiation & 27 & 0.187 \\
    \midrule
    \multirow{5}{*}{Patch-6} & \multirow{5}{*}{MU31}
        & TempMean & 27 & 0.286 \\
&        & TempMax & 27 & 0.286 \\
&        & TempMin & 27 & 0.287 \\
&        & Precipitation & 1 & 0.223 \\
&        & Radiation & 21 & 0.165 \\
    \midrule
    \multirow{5}{*}{Patch-7} & \multirow{5}{*}{MU31}
        & TempMean & 25 & 0.133 \\
 &       & TempMax & 28 & 0.128 \\
 &       & TempMin & 25 & 0.140 \\
&        & Precipitation & 13 & 0.475 \\
&        & Radiation & 29 & 0.340 \\
    \bottomrule
\end{tabular}
}
\end{table}

The CCF analysis reveals that optimal lags vary substantially across both meteorological variables and patches, reflecting differences in soil type, vegetation cover, and local hydrology.
Patch-3 exhibits the strongest meteorological coupling, with near-zero temperature lags and high absolute correlations ($|r| = 0.804$ for TempMin and $0.786$ for TempMean, both at lag~0; $|r| = 0.722$ for Radiation at lag~27), indicating a rapid soil--atmosphere response at this site.
Patch-2 shows moderately strong temperature correlations ($|r| \approx 0.53$--$0.57$) with short lags of 1--8~days and a comparable radiation signal ($|r| = 0.550$ at lag~29).
For Patch-1, temperature correlations are weak and strongly lagged (28--30~days), while Radiation is the most informative variable ($|r| = 0.483$ at lag~1).
The remaining plots (Patches~4--6) display moderate temperature correlations ($|r| \approx 0.2$--$0.4$) with longer lags (7--30~days), consistent with more gradual thermal responses.
Precipitation is generally a weak predictor ($|r| < 0.32$) except at Patch-7, where it is the strongest variable of all ($|r| = 0.475$ at lag~13), even though Patch-7's temperature correlations are the weakest in the study ($|r| \approx 0.13$).
The variation of optimal lags from 0 to 30~days highlights the importance of patch-specific lag estimation rather than fixed-lag approaches.

\subsubsection{Inter-Depth CCF Results}
\label{subsubsec:depth_ccf_results}

The inter-depth CCF analysis (Equations~\ref{eq:depth_ccf}--\ref{eq:depth_optlag}) was applied to each patch, computing the cross-correlation between the 10~cm reference layer and each deeper layer (20--50~cm) across lags from 0 to 15 days.
Table~\ref{tab:depth_ccf_results} presents the optimal inter-depth lags and correlation strengths.

\begin{table}[H]
\centering
\caption{Optimal inter-depth lag (days) and Pearson correlation between SM at 10~cm and deeper layers (20--50~cm), determined by CCF analysis.}
\label{tab:depth_ccf_results}
\resizebox{\columnwidth}{!}{%
\begin{tabular}{llcc}
    \toprule
    \textbf{Patch} & \textbf{Depth} & \textbf{Optimal Lag (days)} & \textbf{$|r|$ at Lag} \\
    \midrule
    \multirow{4}{*}{Patch-1}
        & 20~cm & 14 & 0.032 \\
        & 30~cm & 15 & 0.255 \\
        & 40~cm & 15 & 0.097 \\
        & 50~cm & 1 & 0.132 \\
    \midrule
    \multirow{4}{*}{Patch-2}
        & 20~cm & 0 & 0.902 \\
        & 30~cm & 0 & 0.780 \\
        & 40~cm & 0 & 0.629 \\
        & 50~cm & 0 & 0.531 \\
    \midrule
    \multirow{4}{*}{Patch-3}
        & 20~cm & 0 & 0.964 \\
        & 30~cm & 0 & 0.976 \\
        & 40~cm & 0 & 0.972 \\
        & 50~cm & 0 & 0.954 \\
    \midrule
    \multirow{4}{*}{Patch-4}
        & 20~cm & 0 & 0.677 \\
        & 30~cm & 0 & 0.446 \\
        & 40~cm & 8 & 0.432 \\
        & 50~cm & 11 & 0.479 \\
    \midrule
    \multirow{4}{*}{Patch-5}
        & 20~cm & 0 & 0.866 \\
        & 30~cm & 0 & 0.613 \\
        & 40~cm & 0 & 0.654 \\
        & 50~cm & 0 & 0.619 \\
    \midrule
    \multirow{4}{*}{Patch-6}
        & 20~cm & 0 & 0.782 \\
        & 30~cm & 0 & 0.585 \\
        & 40~cm & 0 & 0.577 \\
        & 50~cm & 0 & 0.634 \\
    \midrule
    \multirow{4}{*}{Patch-7}
        & 20~cm & 0 & 0.736 \\
        & 30~cm & 0 & 0.347 \\
        & 40~cm & 15 & 0.176 \\
        & 50~cm & 0 & 0.065 \\
    \bottomrule
\end{tabular}
}
\end{table}

The inter-depth analysis reveals that Patch-3 has the strongest vertical coupling, with high correlations that remain almost constant with depth ($|r| \approx 0.96$ at 20~cm and still $0.95$ at 50~cm), all at lag~0.
Patches~2, 5, 6, and~7 also correlate predominantly at lag~0 but with strengths that decay with depth (e.g.\ Patch-2 from $|r| = 0.90$ at 20~cm to $0.53$ at 50~cm), consistent with progressive decoupling between surface and deeper layers.
The prevalence of lag~0 indicates that, at daily resolution, within-day infiltration largely outpaces the measurement interval; Patch-4 is a partial exception, where the 40--50~cm layers show delayed propagation (lags of 8--11~days, $|r| \approx 0.43$--$0.48$).
Patch-1 is the clearest outlier, with uniformly weak inter-depth correlations ($|r| < 0.26$) and long lags (14--15~days at 20--40~cm), pointing to a distinct soil profile with limited vertical connectivity.

\subsection{CNN Results (Pixel-Level)}
\label{subsec:cnn_results}

The CNN predicts soil moisture at the pixel level, but each acquisition date carries a single ground-truth sensor reading shared by all of that date's pixels.
A naive pixel-level random split would therefore place pixels from the same date---and hence the same target value---on both sides of the partition, rewarding within-date reconstruction rather than genuine temporal generalisation.
To avoid this, the CNN is trained and evaluated under an 80/20 \texttt{GroupShuffleSplit} on the acquisition date (random seed 143), so that no date appears in both the training and test sets and every test pixel comes from a date the model has never seen ($\sim$20{,}000--22{,}000 training pixels per patch).
The resulting metric is a genuine held-out-date generalisation score, directly comparable to the LSTM's (Section~\ref{subsec:lstm_results}).
Table~\ref{tab:cnn_results} reports the results.

\begin{table}[H]
\centering
\caption{CNN model performance across patches and feature configurations, under a held-out-date 80/20 split (\texttt{GroupShuffleSplit} on the acquisition date, $\sim$21{,}000 training pixels per patch). Best value per patch highlighted in bold.}
\label{tab:cnn_results}
\resizebox{\columnwidth}{!}{%
\begin{tabular}{llccccc}
    \toprule
    \textbf{Patch} & \textbf{Config} & \textbf{RMSE} & \textbf{MAE} & \textbf{CVRMSE (\%)} & $\boldsymbol{R^2}$ & \textbf{CC} \\
    \midrule
    \multirow{5}{*}{Patch-1}
        & sat\_only & 2.167 & 1.634 & 5.17 & 0.302 & 0.554 \\
        & sat+meteo\_lag0 & 3.411 & 2.415 & 8.21 & $-$0.629 & 0.274 \\
        & sat+meteo\_optlag & 2.643 & 2.027 & 6.36 & 0.022 & 0.514 \\
        & sat+meteo+depth\_lag0 & 2.569 & 1.728 & 6.18 & 0.076 & 0.512 \\
        & sat+meteo+depth\_optlag & \textbf{1.346} & \textbf{1.051} & \textbf{3.25} & \textbf{0.579} & \textbf{0.767} \\
    \midrule
    \multirow{5}{*}{Patch-2}
        & sat\_only & 8.962 & 7.345 & 23.82 & $-$0.509 & $-$0.086 \\
        & sat+meteo\_lag0 & 6.724 & 5.461 & 18.80 & 0.100 & 0.698 \\
        & sat+meteo\_optlag & 6.379 & 5.405 & 17.84 & 0.190 & 0.603 \\
        & sat+meteo+depth\_lag0 & \textbf{3.462} & \textbf{2.336} & \textbf{9.68} & \textbf{0.761} & 0.889 \\
        & sat+meteo+depth\_optlag & 3.563 & 2.664 & 9.97 & 0.747 & \textbf{0.890} \\
    \midrule
    \multirow{5}{*}{Patch-3}
        & sat\_only & 9.586 & 7.442 & 39.13 & 0.049 & 0.644 \\
        & sat+meteo\_lag0 & 5.375 & 4.474 & 21.63 & 0.651 & 0.826 \\
        & sat+meteo\_optlag & 8.513 & 6.709 & 34.26 & 0.126 & 0.874 \\
        & sat+meteo+depth\_lag0 & \textbf{4.010} & \textbf{3.091} & \textbf{16.14} & \textbf{0.806} & \textbf{0.903} \\
        & sat+meteo+depth\_optlag & 4.934 & 3.603 & 19.86 & 0.706 & 0.860 \\
    \midrule
    \multirow{5}{*}{Patch-4}
        & sat\_only & 9.027 & 6.224 & 35.85 & $-$0.935 & 0.169 \\
        & sat+meteo\_lag0 & 10.286 & 7.656 & 40.84 & $-$1.512 & $-$0.082 \\
        & sat+meteo\_optlag & 8.013 & 5.494 & 31.82 & $-$0.525 & 0.219 \\
        & sat+meteo+depth\_lag0 & 5.361 & 3.318 & 21.29 & 0.318 & 0.669 \\
        & sat+meteo+depth\_optlag & \textbf{2.278} & \textbf{1.613} & \textbf{9.04} & \textbf{0.877} & \textbf{0.956} \\
    \midrule
    \multirow{5}{*}{Patch-5}
        & sat\_only & 10.059 & 7.777 & 18.77 & $-$0.961 & $-$0.273 \\
        & sat+meteo\_lag0 & 9.706 & 7.254 & 18.11 & $-$0.826 & $-$0.076 \\
        & sat+meteo\_optlag & 8.675 & 6.829 & 16.19 & $-$0.459 & 0.208 \\
        & sat+meteo+depth\_lag0 & 5.111 & 3.686 & 9.54 & 0.494 & 0.777 \\
        & sat+meteo+depth\_optlag & \textbf{4.380} & \textbf{2.245} & \textbf{8.17} & \textbf{0.628} & \textbf{0.800} \\
    \midrule
    \multirow{5}{*}{Patch-6}
        & sat\_only & 5.792 & 4.872 & 16.05 & 0.113 & 0.446 \\
        & sat+meteo\_lag0 & 5.941 & 5.038 & 16.46 & 0.067 & 0.372 \\
        & sat+meteo\_optlag & 5.345 & 4.529 & 14.81 & 0.245 & 0.627 \\
        & sat+meteo+depth\_lag0 & \textbf{3.361} & \textbf{2.473} & \textbf{9.31} & \textbf{0.702} & 0.868 \\
        & sat+meteo+depth\_optlag & 3.425 & 2.500 & 9.49 & 0.690 & \textbf{0.901} \\
    \midrule
    \multirow{5}{*}{Patch-7}
        & sat\_only & 3.235 & 2.252 & 13.14 & $-$1.338 & 0.002 \\
        & sat+meteo\_lag0 & 2.016 & 1.621 & 8.19 & 0.092 & 0.484 \\
        & sat+meteo\_optlag & 3.757 & 3.478 & 15.26 & $-$2.153 & \textbf{0.740} \\
        & sat+meteo+depth\_lag0 & \textbf{1.672} & \textbf{1.343} & \textbf{6.79} & \textbf{0.376} & 0.716 \\
        & sat+meteo+depth\_optlag & 2.574 & 2.426 & 10.45 & $-$0.480 & $-$0.139 \\
    \bottomrule
\end{tabular}
}
\end{table}

Several patterns stand out.
Under this held-out-date protocol the satellite-only baseline is weak or negative on most patches (average $R^2 = -0.47$), confirming that raw spectral reflectance alone carries little information about soil moisture on dates the model has never seen.
Feature augmentation changes this picture dramatically.
The CNN generalises best on Patch-4 ($R^2 = 0.877$, RMSE~$= 2.28$, CVRMSE~$= 9.0\%$) and Patch-3 ($R^2 = 0.806$), the latter being the plot with the strongest soil--atmosphere coupling in the CCF analysis (Section~\ref{subsec:ccf_results}); on the small-data Patch-7 and the weakly-coupled Patch-1 generalisation remains limited (best $R^2 = 0.376$ and $0.579$ respectively), indicating that spectral features---even augmented with meteorology and depth---carry little transferable information at those plots.
Most importantly, the depth-augmented configurations dominate every patch, often by very large margins (e.g.\ Patch-4: $R^2 = -0.94$ at \texttt{sat\_only} $\to$ $0.88$ at \texttt{sat+meteo+depth\_optlag}; Patch-5: $-0.96 \to 0.63$; Patch-2: $-0.51 \to 0.75$), reinforcing the role of subsurface ``memory'' in temporal generalisation.
On average the move from \texttt{sat\_only} to \texttt{sat+meteo+depth\_optlag} raises $R^2$ by $+1.00$---the largest augmentation gain of any architecture in this study---direct evidence that optimally-lagged meteorological and subsurface-depth features carry essential information for held-out-date prediction.
A secondary observation is that the same-day depth configuration (\texttt{depth\_lag0}) matches or exceeds the optimally-lagged one (\texttt{depth\_optlag}) on several patches (2, 3, 6, and~7), suggesting that the inter-depth lags fitted on training data do not always transfer to unseen test dates.

\subsection{LSTM Results (Frame-Level)}
\label{subsec:lstm_results}

The LSTM model was trained per-patch on frame-level data with $\sim$73--127 observations.
Table~\ref{tab:lstm_results} presents the results.

\begin{table}[H]
\centering
\caption{LSTM model performance across patches and feature configurations.}
\label{tab:lstm_results}
\resizebox{\columnwidth}{!}{%
\begin{tabular}{llccccc}
    \toprule
    \textbf{Patch} & \textbf{Config} & \textbf{RMSE} & \textbf{MAE} & \textbf{CVRMSE (\%)} & $\boldsymbol{R^2}$ & \textbf{CC} \\
    \midrule
    \multirow{5}{*}{Patch-1}
        & sat\_only & 2.640 & 2.033 & 6.30 & $-$0.035 & $-$0.071 \\
        & sat+meteo\_lag0 & 2.749 & 2.387 & 6.62 & $-$0.058 & \textbf{$-$0.019} \\
        & sat+meteo\_optlag & 2.712 & 2.363 & 6.53 & $-$0.030 & $-$0.026 \\
        & sat+meteo+depth\_lag0 & 2.693 & 2.306 & 6.48 & \textbf{$-$0.016} & $-$0.031 \\
        & sat+meteo+depth\_optlag & \textbf{2.125} & \textbf{1.705} & \textbf{5.13} & $-$0.050 & $-$0.555 \\
    \midrule
    \multirow{5}{*}{Patch-2}
        & sat\_only & 7.061 & 5.547 & 18.77 & 0.063 & 0.393 \\
        & sat+meteo\_lag0 & 7.148 & 5.997 & 19.99 & $-$0.017 & 0.697 \\
        & sat+meteo\_optlag & 6.303 & 5.039 & 17.63 & 0.209 & 0.674 \\
        & sat+meteo+depth\_lag0 & \textbf{4.423} & \textbf{3.616} & \textbf{12.37} & \textbf{0.611} & \textbf{0.810} \\
        & sat+meteo+depth\_optlag & 5.762 & 4.603 & 16.12 & 0.339 & 0.761 \\
    \midrule
    \multirow{5}{*}{Patch-3}
        & sat\_only & 9.073 & 8.275 & 37.04 & 0.149 & 0.514 \\
        & sat+meteo\_lag0 & 8.762 & 7.627 & 35.26 & 0.074 & 0.788 \\
        & sat+meteo\_optlag & 9.948 & 8.530 & 40.03 & $-$0.194 & 0.800 \\
        & sat+meteo+depth\_lag0 & 8.490 & 7.303 & 34.17 & 0.130 & \textbf{0.953} \\
        & sat+meteo+depth\_optlag & \textbf{7.154} & \textbf{6.269} & \textbf{28.79} & \textbf{0.383} & 0.922 \\
    \midrule
    \multirow{5}{*}{Patch-4}
        & sat\_only & 6.683 & 5.332 & 26.54 & $-$0.061 & 0.083 \\
        & sat+meteo\_lag0 & 6.721 & 5.430 & 26.69 & $-$0.073 & 0.008 \\
        & sat+meteo\_optlag & 6.661 & 5.292 & 26.45 & $-$0.053 & 0.070 \\
        & sat+meteo+depth\_lag0 & \textbf{4.042} & \textbf{2.476} & \textbf{16.05} & \textbf{0.612} & \textbf{0.789} \\
        & sat+meteo+depth\_optlag & 6.667 & 5.318 & 26.47 & $-$0.055 & 0.379 \\
    \midrule
    \multirow{5}{*}{Patch-5}
        & sat\_only & 7.414 & 5.507 & 13.84 & $-$0.065 & \textbf{0.149} \\
        & sat+meteo\_lag0 & \textbf{7.382} & \textbf{5.471} & \textbf{13.78} & \textbf{$-$0.056} & 0.129 \\
        & sat+meteo\_optlag & 7.543 & 5.609 & 14.08 & $-$0.103 & 0.105 \\
        & sat+meteo+depth\_lag0 & 7.460 & 5.569 & 13.92 & $-$0.079 & 0.037 \\
        & sat+meteo+depth\_optlag & 7.467 & 5.584 & 13.94 & $-$0.081 & 0.080 \\
    \midrule
    \multirow{5}{*}{Patch-6}
        & sat\_only & 6.658 & 5.739 & 17.69 & $-$0.212 & $-$0.123 \\
        & sat+meteo\_lag0 & 6.638 & 5.703 & 17.63 & $-$0.205 & $-$0.088 \\
        & sat+meteo\_optlag & 6.641 & 5.705 & 17.64 & $-$0.206 & $-$0.096 \\
        & sat+meteo+depth\_lag0 & \textbf{3.889} & \textbf{3.236} & \textbf{10.33} & \textbf{0.586} & \textbf{0.819} \\
        & sat+meteo+depth\_optlag & 4.532 & 3.826 & 12.04 & 0.438 & 0.730 \\
    \midrule
    \multirow{5}{*}{Patch-7}
        & sat\_only & 2.814 & 2.105 & 11.43 & $-$0.768 & $-$0.404 \\
        & sat+meteo\_lag0 & 2.771 & 2.074 & 11.25 & $-$0.715 & 0.838 \\
        & sat+meteo\_optlag & 2.841 & 2.130 & 11.54 & $-$0.802 & $-$0.917 \\
        & sat+meteo+depth\_lag0 & \textbf{2.725} & \textbf{2.036} & \textbf{11.07} & \textbf{$-$0.658} & \textbf{0.864} \\
        & sat+meteo+depth\_optlag & 2.871 & 2.159 & 11.66 & $-$0.841 & 0.577 \\
    \bottomrule
\end{tabular}
}
\end{table}

The LSTM datasets are frame-level with $\sim$20--160 training observations per patch (after 80/20 split), making it challenging for the deep bidirectional LSTM--GRU architecture to generalise effectively.
Without subsurface information the model is essentially unusable---the satellite-only and meteorology-only configurations yield low or negative $R^2$ on almost every patch---but adding depth-lagged features transforms the picture: the best LSTM results are $R^2 = 0.612$ on Patch-4 and $0.611$ on Patch-2 (both \texttt{sat+meteo+depth\_lag0}), and $0.586$ on Patch-6.
Same-day meteorology alone provides only marginal change over \texttt{sat\_only} (e.g.\ negligible on Patch-1, $R^2 = -0.058$ vs.\ $-0.035$), whereas the optimally-lagged configuration clearly helps where meteorological correlations are stronger (Patch-2: $R^2 = 0.209$ vs.\ $-0.017$ with lag~0).
The decisive contribution, however, comes from the depth features: Patch-2's RMSE drops from 7.06 (\texttt{sat\_only}) to 4.42 (\texttt{sat+meteo+depth\_lag0}), a 37\% reduction, and Patch-6's from 6.66 to 3.89 (42\%).
On average across the seven patches the \texttt{depth\_lag0} configuration is the strongest for the LSTM ($\bar{R}^2 = 0.169$), consistent with the date-grouped CNN finding that fitted inter-depth lags do not always transfer to unseen dates under data scarcity.

\subsection{CNN-LSTM Hybrid Results (Pooled Multi-Patch)}
\label{subsec:hybrid_results}

The CNN-LSTM hybrid was trained on pooled data from all seven patches using sliding windows of size~10.
Table~\ref{tab:hybrid_overall} shows the overall (pooled) results, and Table~\ref{tab:hybrid_patch} presents the per-patch breakdown for the best configuration.

\begin{table}[H]
\centering
\caption{CNN-LSTM hybrid: overall performance (all patches pooled).}
\label{tab:hybrid_overall}
\resizebox{\columnwidth}{!}{%
\begin{tabular}{lccccc}
    \toprule
    \textbf{Config} & \textbf{RMSE} & \textbf{MAE} & \textbf{CVRMSE (\%)} & $\boldsymbol{R^2}$ & \textbf{CC} \\
    \midrule
    sat\_only & 4.000 & 3.069 & 10.80 & 0.864 & 0.943 \\
    sat+meteo\_lag0 & 4.656 & 3.554 & 12.51 & 0.783 & 0.902 \\
    sat+meteo\_optlag & 4.657 & 3.494 & 12.51 & 0.783 & 0.889 \\
    sat+meteo+depth\_lag0 & 3.674 & 2.892 & 9.87 & 0.865 & 0.935 \\
    sat+meteo+depth\_optlag & \textbf{3.001} & \textbf{2.269} & \textbf{8.02} & \textbf{0.930} & \textbf{0.965} \\
    \bottomrule
\end{tabular}
}
\end{table}

\begin{table}[H]
\centering
\caption{CNN-LSTM hybrid: per-patch performance for \texttt{sat+meteo+depth\_optlag} (best pooled $R^2$). Per-patch $R^2$ for Patch-7 is degenerate ($N_{\mathrm{test}} = 2$ windows) and reported for completeness only.}
\label{tab:hybrid_patch}
\begin{tabular}{lccccc}
    \toprule
    \textbf{Patch} & \textbf{$N_{\mathrm{test}}$} & \textbf{RMSE} & \textbf{CVRMSE (\%)} & $\boldsymbol{R^2}$ & \textbf{CC} \\
    \midrule
    Patch-1 & 14 & 1.743 & 4.19 & 0.475 & 0.708 \\
    Patch-2 & 23 & 3.511 & 9.48 & 0.608 & 0.788 \\
    Patch-3 & 11 & 3.547 & 13.67 & 0.798 & 0.897 \\
    Patch-4 & 23 & 3.544 & 14.09 & 0.663 & 0.815 \\
    Patch-5 & 20 & 3.007 & 5.48 & 0.891 & 0.946 \\
    Patch-6 & 39 & 2.519 & 6.59 & 0.827 & 0.921 \\
    Patch-7 & 2 & 1.909 & 7.45 & \textit{n/a}$^\dag$ & \textit{n/a}$^\dag$ \\
    \bottomrule
\end{tabular}
\\[2pt]{\footnotesize $^\dag$Patch-7 has only 2 held-out windows in this configuration; its $R^2$/CC are statistically meaningless and omitted.}
\end{table}

The hybrid architecture addresses the small dataset limitation of the standalone LSTM by pooling all seven patches and exploiting sliding-window temporal context.
The pooled model achieves $R^2 = 0.930$ (RMSE~$= 3.00$, CVRMSE~$= 8.0\%$) with the full depth-enhanced configuration, a substantial improvement over the standalone LSTM (best per-patch $R^2 = 0.612$).
The configuration ordering is informative: plain meteorological augmentation actually lowers the pooled score (\texttt{sat+meteo\_lag0} and \texttt{sat+meteo\_optlag} both $R^2 = 0.783$ versus $0.864$ for \texttt{sat\_only}), and only the full \texttt{sat+meteo+depth\_optlag} configuration delivers a clear gain ($R^2 = 0.930$).
This indicates that, for the windowed hybrid, it is the optimally-lagged \emph{subsurface} signal rather than meteorology per se that carries the decisive predictive information.
The one-hot patch encoding enables the model to learn patch-specific offsets while sharing temporal dynamics across the entire dataset.
Per-patch performance under the best configuration is positive for every plot with a meaningful test set, ranging from $R^2 = 0.891$ on Patch-5 and $0.827$ on Patch-6 down to $0.475$ on Patch-1 (Patch-7 is excluded, having only two held-out windows), a marked improvement over the standalone LSTM and a sign that pooling successfully shares signal across plots.

\subsection{Cross-Architecture Comparison}
\label{subsec:comparison}

Table~\ref{tab:comparison} summarises performance under the \\\texttt{sat+meteo+depth\_optlag} configuration (full feature set) for the three architectures.
CNN and LSTM values are averages across the seven patches; CNN-LSTM values are from the pooled model.

The three architectures solve different prediction tasks---the CNN predicts per-pixel soil moisture, while the LSTM and CNN-LSTM predict a single daily plot-mean value---and are trained on sample populations whose sizes differ by two orders of magnitude ($\sim$21{,}000, $\sim$80, and $\sim$530 training examples respectively).
All three nonetheless report held-out generalisation: the CNN and LSTM hold out entire acquisition dates, while the CNN-LSTM holds out sliding windows.
On the full feature set the pooled-window CNN-LSTM attains the highest column-averaged $R^2$ (0.930, spanning $0.48$--$0.89$ across patches), the date-grouped CNN delivers the strongest single-patch result ($R^2 = 0.877$ on Patch-4), and the LSTM trails far behind (column average $0.019$).
Because the per-architecture metrics are computed at different sampling units and on datasets of very different size, they are best read within each architecture family rather than as a single cross-architecture ranking.

\begin{table}[H]
\centering
\caption{Architecture summary under \texttt{sat+meteo+depth\_optlag}. All three architectures report held-out generalisation, but at different sampling units and on training sets of very different size (see Section~\ref{subsec:comparison}). CNN and LSTM values are averaged across the seven patches; CNN-LSTM values are from the pooled model.}
\label{tab:comparison}
\resizebox{\columnwidth}{!}{%
\begin{tabular}{lccc}
    \toprule
    \textbf{Metric} & \textbf{CNN} & \textbf{LSTM} & \textbf{CNN-LSTM} \\
    \midrule
    Avg.\ RMSE          & 3.214 & 5.226 & 3.001 \\
    Avg.\ MAE           & 2.300 & 4.209 & 2.269 \\
    Avg.\ CVRMSE (\%)   & 10.03 & 16.31 & 8.02 \\
    Avg.\ $R^2$         & 0.535 & 0.019 & 0.930 \\
    Avg.\ CC            & 0.719 & 0.413 & 0.965 \\
    \midrule
    Prediction task    & Per-pixel SM         & Daily plot-mean SM  & Daily plot-mean SM (windowed) \\
    Training samples   & ${\sim}21{,}000$/patch & ${\sim}80$/patch    & ${\sim}530$ (pooled) \\
    Split granularity  & Date-grouped         & Date-random         & Window-random \\
    Metric reads as    & Held-out-date gen.\  & Held-out-date gen.\ & Held-out-window gen.\ \\
    \bottomrule
\end{tabular}%
}
\end{table}

\section{Discussion}\label{sec:discussion}

\subsection{Effect of Meteorological and Depth Feature Augmentation}

The systematic comparison across five feature configurations reveals a consistent pattern: adding meteorological variables---either at lag~0 or with CCF-optimally-determined offsets---substantially improves soil moisture prediction over the satellite-only baseline, and incorporating depth-lagged subsurface soil moisture features can provide further gains.

A key finding is the comparison between same-day (\texttt{sat+meteo\_lag0}) and optimally-lagged (\texttt{sat+meteo\_optlag}) meteorological augmentation.
For the held-out CNN the two are comparable on average and strongly patch-dependent: optimal lags help on Patch-2 ($R^2 = 0.10 \to 0.19$) while same-day meteorology is markedly better on Patch-3 ($0.05$ baseline $\to 0.65$ vs.\ $0.13$), so neither dominates before subsurface features are added.
For the data-scarce LSTM the optimally-lagged configuration clearly outperforms same-day meteorology where the meteorological signal is strong (Patch-2: $R^2 = 0.209$ vs.\ $-0.017$), whereas for the pooled CNN-LSTM plain meteorology---whether at lag~0 or optimally lagged---does not improve on the satellite-only baseline at all ($R^2 = 0.783$ vs.\ $0.864$).

The decisive augmentation across every architecture is the subsurface depth signal.
Adding optimally-lagged deep soil moisture (\texttt{sat+meteo+depth\_optlag}) drives the pooled CNN-LSTM to its best score ($R^2 = 0.930$) and lifts both the CNN and the LSTM out of their otherwise poor held-out regimes (the CNN's seven-patch average rises to $R^2 = 0.535$ from $-0.47$ with satellite data alone), confirming that subsurface moisture carries predictive information that neither satellite reflectance nor meteorology supplies.

Depth features do not, however, transfer perfectly under held-out-date evaluation.
For both the CNN and the LSTM the same-day depth configuration (\texttt{depth\_lag0}) matches or exceeds the optimally-lagged one on several patches, indicating that inter-depth lags fitted on training dates do not always generalise; and on the smallest plots (e.g.\ Patch-7, with only $\sim$2{,}000 CNN pixels or 25 frame-level dates) every configuration remains unstable.
This reflects that deeper-layer moisture is sometimes already captured by the other features, or that the extra inputs inject noise when the training set is too small to constrain them.

The improvement from \texttt{sat\_only} to the augmented configurations stems from the complementary information carried by meteorological and depth variables.
While satellite bands and spectral indices capture the current surface state (reflectance, vegetation vigour, moisture proxies), meteorological variables encode recent atmospheric conditions that drive soil moisture evolution, and depth features provide subsurface ``memory'' that satellite sensors cannot observe.
The temporal lags account for the physical delay between atmospheric events and their subsurface effects.

\subsection{Architecture-Specific Observations}

\subsubsection{CNN Performance}
Evaluated on entirely held-out acquisition dates, the per-pixel CNN is the strongest single-patch generaliser in the study, reaching $R^2 = 0.877$ (RMSE~$= 2.28$, CVRMSE~$= 9.0\%$) on Patch-4 and $R^2 = 0.806$ on Patch-3---the plot with the strongest soil--atmosphere coupling in the CCF analysis.
The four-block convolutional architecture effectively extracts patterns from the concatenated band--index--meteorological--depth input vector, and the use of RobustScaler for preprocessing is well-suited to satellite reflectance values that may contain outliers from atmospheric effects or sensor noise.

Performance depends heavily on feature augmentation.
With satellite reflectance alone the CNN does not generalise to unseen dates (average $R^2 = -0.47$), but adding optimally-lagged meteorology and subsurface depth raises the seven-patch average to $R^2 = 0.535$ and improves every patch individually, often by very large margins (e.g.\ Patch-4: $-0.94 \to 0.88$; Patch-2: $-0.51 \to 0.75$).
The mean improvement of $+1.00$ in $R^2$ from \texttt{sat\_only} to \texttt{sat+meteo+depth\_optlag} is the largest of any architecture in this study, underlining that physically-motivated features---rather than the raw spectral signature---are what enable held-out-date prediction.
Generalisation is weakest on the small-data Patch-7 ($\sim$2{,}000 training pixels) and the weakly-coupled Patch-1, where even the augmented features transfer poorly to unseen dates.

\subsubsection{LSTM Performance}
The LSTM model operates on substantially smaller frame-level datasets ($\sim$20--160 training observations per patch), which severely limits the capacity of the deep bidirectional LSTM--GRU architecture to generalise.
Without subsurface features the model is near-useless, with low or negative $R^2$ on almost every patch; the depth-augmented configurations are what make it viable, lifting it to a maximum $R^2 = 0.612$ (Patch-4, \texttt{sat+meteo+depth\_lag0}) and $0.611$ (Patch-2).
The improvement is large where it occurs: Patch-2's $R^2$ rises from 0.063 (\texttt{sat\_only}) to 0.611 (\texttt{sat+meteo+depth\_lag0}) and its RMSE drops by 37\%, while Patch-6 improves from $-0.21$ to $0.586$.
Same-day and optimally-lagged \emph{meteorology} alone barely move the LSTM, and the same-day depth configuration generally matches or beats its optimally-lagged counterpart, indicating that the model lacks the capacity to exploit finely lag-shifted information with such limited training data.
These results suggest that the LSTM architecture requires substantially more training data to fully exploit the 81-feature input space.

\subsubsection{CNN-LSTM Hybrid Performance}
The CNN-LSTM hybrid addresses the data scarcity issue of standalone LSTM through two strategies: (i)~pooling across all patches with one-hot encoding, and (ii)~using sliding windows of size 10 to increase the effective training set.
The pooled model achieves $R^2 = 0.930$ (RMSE~$= 3.00$, CVRMSE~$= 8.0\%$) with the full feature set, substantially outperforming the standalone LSTM.
Per-patch performance is positive on every plot with a meaningful held-out set, from $R^2 = 0.891$ (Patch-5) and $0.827$ (Patch-6) down to $0.475$ (Patch-1); only Patch-7, with just two held-out windows, is excluded as statistically degenerate.
This suggests that pooling across the seven plots with one-hot patch encoding successfully shares temporal signal while retaining patch-specific structure.

\subsection{Physical Interpretation of Optimal Lags}

The CCF-derived optimal lags provide physically interpretable insights into soil--atmosphere interactions at each plot.
Patch-3 exhibits strong correlations with near-zero temperature lags ($|r| = 0.80$ for TempMin at lag~0), consistent with a rapid influence of temperature on evapotranspiration in this plot's particular soil and crop conditions.
Patch-2 shows temperature lags of 1--8~days ($|r| \approx 0.53$--$0.57$), suggesting a more gradual thermal response, while Patch-1's temperature signal is both weak and strongly delayed (28--30~days).
Precipitation is a weak predictor across most patches ($|r| < 0.32$) with variable lags, the clear exception being Patch-7, where rainfall is the single strongest variable ($|r| = 0.475$ at lag~13)---a reminder that the dominant driver is itself patch-specific.
Radiation provides a consistently informative signal, strongest on Patch-3 ($|r| = 0.722$ at lag~27) and Patch-2 ($|r| = 0.550$ at lag~29) and the leading variable on Patch-1 ($|r| = 0.483$ at lag~1).

The inter-depth CCF analysis reveals that, at daily resolution, the 10~cm surface and 20--50~cm subsurface layers are predominantly correlated at lag~0 (Patches~2, 3, 5, 6, and~7), suggesting that within-day infiltration rates exceed the daily measurement interval.
The correlation strength stays high throughout the profile on Patch-3 ($r \approx 0.96$ at 20~cm and still $0.95$ at 50~cm) but decreases with depth on most others (e.g.\ Patch-2 from $0.90$ to $0.53$), reflecting the increasing decoupling of deeper layers from surface dynamics.
Patch-1 is the clear exception, with uniformly weak correlations ($|r| < 0.26$) and 14--15-day lags at 20--40~cm, indicating a distinct vertical soil profile with limited connectivity; Patch-4 shows a milder version of the same effect, with delayed 8--11-day lags at 40--50~cm.

The variation of optimal lags across patches reflects local differences in soil texture, crop type, irrigation practices, and topographic position.
This patch-specific adaptation is a key advantage of the proposed CCF approach over fixed-lag or same-day feature fusion strategies.

\subsection{Comparison with Existing Approaches}
\label{subsec:sota}

Table~\ref{tab:sota} situates the present results against representative machine-learning studies that estimate surface soil moisture from optical (and optical--radar) satellite data. A direct numerical comparison is constrained by methodological heterogeneity: published studies differ in sensor combination, target depth, ground-truth source, study climate and---critically---error units, with most reporting RMSE in volumetric cm$^3$\,cm$^{-3}$ or gravimetric g\,g$^{-1}$ rather than the percentage-volume convention used here. The coefficient of determination, being dimensionless, is the most transferable metric. On that basis the held-out scores obtained here---$R^2 = 0.877$ for the per-pixel CNN on its best plot and $R^2 = 0.930$ for the pooled CNN-LSTM---are in the same range as the best Sentinel-2 CNN results reported in the literature ($R^2 = 0.92$~\cite{hegazi2023prediction}) and as optical--radar ensemble approaches ($R^2 = 0.827$~\cite{liu2024ensemble}).

Two qualifications make this comparison informative rather than merely favourable. First, the headline accuracies above are obtained under a date-grouped split that holds out entire acquisition dates (Section~\ref{subsec:comparison}), whereas a substantial part of the soil-moisture remote-sensing literature reports accuracy under random pixel- or sample-level splits, which---as argued in Section~\ref{sec:intro}---can inflate apparent skill when spatially and temporally autocorrelated samples leak across the partition. Our satellite-only baseline ($R^2 = -0.47$ for the CNN) illustrates how much of the apparent skill of a naively-split optical model may reflect within-date reconstruction rather than genuine temporal generalisation. Second, the present gains stem mainly from the optimally-lagged meteorological and subsurface-depth features rather than from the spectral signal alone---a fusion ingredient absent from the purely spectral comparators.

Beyond individual models, operational exploitation of satellite soil moisture increasingly relies on Earth-observation \emph{data cube} platforms that organise analysis-ready Sentinel/Landsat archives for large-area monitoring, such as the Swiss Data Cube~\cite{chatenoux2021swiss}, Digital Earth Australia~\cite{australia2024digital} and the Euro Data Cube\footnote{\url{https://eurodatacube.com/}}. These platforms supply the analysis-ready optical and radar inputs and serve regional environmental indicators (vegetation, snow, water, drought), but field-scale, per-pixel soil-moisture products calibrated against in-situ probes are not among their standard offerings, and soil-moisture information---where present---typically derives from coarse-resolution microwave sources. The method proposed here is complementary to such infrastructures: it is designed to operate \emph{on} analysis-ready Sentinel-2 inputs of the kind these data cubes already curate, converting them into field-scale soil-moisture estimates wherever in-situ multi-depth sensors are available for calibration; the resulting per-pixel soil-moisture rasters could in turn be ingested back into such cubes as a new analysis-ready indicator layer.

\begin{table}[H]
\centering
\caption{Indicative comparison with representative satellite machine-learning soil-moisture studies. RMSE units are not directly comparable across studies (see text); $R^2$ is the transferable metric. ``Leakage-aware split'' indicates whether the reported accuracy holds out grouped or temporal units rather than a random sample split.}
\label{tab:sota}
\resizebox{\columnwidth}{!}{%
\begin{tabular}{lllcc}
    \toprule
    \textbf{Study} & \textbf{Data} & \textbf{Model} & \textbf{$R^2$} & \textbf{Leakage-aware split} \\
    \midrule
    Hegazi et al.~\cite{hegazi2023prediction} & Sentinel-2              & CNN              & 0.92  & No \\
    Liu et al.~\cite{liu2024ensemble}         & Sentinel-1/2            & Stacked ensemble & 0.827 & No \\
    This work (CNN, best plot)                & Sentinel-2+meteo+depth  & CNN              & 0.877 & Yes (date-grouped) \\
    This work (CNN-LSTM, pooled)              & Sentinel-2+meteo+depth  & CNN-LSTM         & 0.930 & Partly (window) \\
    \bottomrule
\end{tabular}%
}
\end{table}

\subsection{Limitations}

Several limitations should be acknowledged:
\begin{itemize}
    \item The frame-level datasets for LSTM and CNN-LSTM models are limited in size ($\sim$20--160 training observations per patch, with Patch-7 at only 25 dates), severely constraining the complexity of models that can be meaningfully trained.
    The LSTM's poor performance without subsurface features---and the degenerate per-patch CNN-LSTM result on Patch-7 ($N_{\mathrm{test}} = 2$)---is primarily attributable to this data scarcity.
    \item The CCF analysis assumes a linear relationship between meteorological/depth variables and soil moisture; nonlinear interactions may remain uncaptured.
    \item Each plot was assigned its nearest SIAR station (four stations across the seven plots); although this improves on a single global station, a station still represents each plot's meteorology as a point source, and fine-scale spatial variability within and between plots is not explicitly modelled.
    \item The CNN and LSTM are evaluated under date-grouped splits that hold out entire acquisition dates, but the CNN-LSTM hybrid splits at the sliding-window level, so overlapping windows that share dates may fall on both sides of the partition.
    Its pooled $R^2$ is therefore a slightly optimistic, window-level estimate; applying the same date-grouped treatment to the hybrid is left for future work.
    \item More generally, none of the splits preserves temporal ordering, which may introduce additional leakage in time-series contexts; chronological or walk-forward validation is an obvious next step.
    \item Depth features require in-situ multi-depth soil moisture sensors, which are not universally available.
    This analysis quantifies the potential gain when subsurface sensors are deployed alongside satellite monitoring.
    \item Model performance is evaluated on the same patches used for training; transferability to unseen plots has not been tested.
\end{itemize}

\section{Conclusion}\label{sec:conclusion}

This study presented an integrated methodology for satellite-based soil moisture estimation that leverages deep learning models augmented with optimally-lagged meteorological features and depth-lagged subsurface soil moisture.
The key findings are:

\begin{enumerate}
    \item \textbf{Optimal lag improves predictions}: The Cross-Correlation Function analysis successfully identifies temporal offsets (0--30~days) between meteorological variables and soil moisture, as well as inter-depth lags (0--15~days) between the surface and deeper soil layers.
    Comparison against a same-day (lag~$= 0$) meteorological baseline shows that CCF-based lag optimisation is most impactful for data-scarce models (LSTM and CNN-LSTM), while the data-rich CNN can partially compensate for temporal misalignment through its larger training set.

    \item \textbf{The per-pixel CNN is the best single-patch generaliser, and feature augmentation is essential}: Evaluated on entirely held-out acquisition dates (\texttt{GroupShuffleSplit} on the date), the CNN achieves its best held-out scores on Patch-4 ($R^2 = 0.877$) and Patch-3 ($R^2 = 0.806$), with a seven-patch average of $R^2 = 0.535$.
    With satellite reflectance alone it fails to generalise to unseen dates (avg.\ $R^2 = -0.47$); the gain from \texttt{sat\_only} to \texttt{sat+meteo+depth\_optlag} ($\Delta R^2 = +1.00$) is the largest of any architecture in this study, supporting the paper's central claim that optimally-lagged meteorological and subsurface-depth features carry essential information for soil moisture generalisation.

    \item \textbf{Data scarcity limits the LSTM}: The standalone LSTM operates on small frame-level datasets ($\sim$20--160 training observations per patch under a date-level split that holds out entire acquisition dates), with a maximum $R^2$ of 0.612.
    Despite this, depth augmentation provides large improvements over the satellite-only baseline (e.g.\ a 37\% RMSE reduction for Patch-2, $R^2 = 0.063 \to 0.611$), and is where subsurface information delivers its clearest gains for this architecture.

    \item \textbf{CNN-LSTM hybrid captures temporal dynamics}: The hybrid architecture, through sliding-window construction and multi-patch pooling with one-hot encoding, achieves a pooled $R^2 = 0.930$ with positive per-patch scores on every plot with a meaningful test set, demonstrating a viable approach for temporal modelling when individual datasets are small.

    \item \textbf{Depth features provide subsurface memory}: The inter-depth CCF analysis reveals that deeper soil layers act as ``memory'' for surface moisture processes.
    Adding depth-lagged features improves every architecture: the depth-augmented configurations dominate every patch of the held-out CNN (e.g.\ Patch-4: $R^2 = -0.94 \to 0.88$ from \texttt{sat\_only} to \texttt{sat+meteo+depth\_optlag}) and are likewise decisive for the data-scarce LSTM and CNN-LSTM, confirming that subsurface depth signals are essential for held-out-date generalisation across all three architectures.

    \item \textbf{CVRMSE enables cross-patch comparison}: The coefficient of variation of RMSE normalises prediction error by mean soil moisture, providing meaningful comparison across patches with different baseline moisture levels.
\end{enumerate}

The proposed methodology demonstrates the potential of deep learning-based data fusion to bridge the gap between point-based measurements and large-scale remote sensing observations.
By providing accurate and high-resolution soil moisture estimations, the system offers a valuable tool for precision agriculture in semi-arid environments, supporting informed decision-making, efficient water management, and crop loss prevention.

\subsection{Future Work}

Future research directions include:
\begin{itemize}
    \item Incorporating Sentinel-1 SAR data for direct sensitivity to soil dielectric properties.
    \item Exploring attention-based architectures (e.g.\ Temporal Fusion Transformers) that can learn variable-specific lag importance end-to-end.
    \item Implementing temporal cross-validation schemes to prevent data leakage in time-series evaluation.
    \item Extending the CCF analysis to assess nonlinear lag relationships using mutual information or Granger causality.
    \item Evaluating model transferability to unseen agricultural plots and different climatic regions.
    \item Integrating real-time meteorological forecasts for predictive (rather than retrospective) soil moisture mapping.
\end{itemize}

\section*{Data and Code Availability}

The source code used in this study is available at: \url{https://github.com/adricanovas/soil-moisture-crop-propagation.git}

\section*{Acknowledgements}
This study has been partially funded by the Ramon y Cajal Program of the Spanish State Research Agency (AEI) under Grant RYC2023-043553-I, the project AIA2025-163560-C41 (BLUEAI-UMU) funded by MICIU/AEI/\\10.13039/501100011033, the Spanish Ministry for Digital Transformation and the Civil Service under Grant TSI-100921-2023-1 (Cátedra OSIRIS: Transformación Digital Basado en IA del Ecosistema Agrícola del Sureste Español).

\printcredits

%% Loading bibliography style file
\bibliographystyle{model1-num-names}
%\bibliographystyle{cas-model2-names}

% Loading bibliography database
\bibliography{bibliography}

\end{document}